\documentclass[sigconf]{acmart}
\AtBeginDocument{%
  }
\usepackage{multirow}
\usepackage{algorithm}
\usepackage{algorithmic}

\copyrightyear{2025}
\acmYear{2025}
\setcopyright{cc}
\setcctype{by-nc-sa}
\acmConference[ICAIF '25]{6th ACM International Conference on AI in
Finance}{November 15--18, 2025}{Singapore, Singapore}
\acmBooktitle{6th ACM International Conference on AI in Finance (ICAIF '25),
November 15--18, 2025, Singapore,
Singapore}\acmDOI{10.1145/3768292.3770414}
\acmISBN{979-8-4007-2220-2/2025/11}

\begin{document}

\title{From News to Returns: A Granger-Causal Hypergraph Transformer on the Sphere}

% \affiliation{%

% \affiliation{%

% \affiliation{%

\author{Anoushka Harit}
\affiliation{%
  \institution{University of Cambridge}
  \country{United Kingdom}
}
\email{ah2415@cam.ac.uk}

\author{Zhongtian Sun}
\affiliation{%
  \institution{University of Kent}
  \institution{Department of Computer Science, University of Cambridge}
  \country{United Kingdom}
}
\email{zs256@kent.ac.uk}

\author{Jongmin Yu}
\affiliation{%
  \institution{ProjectG.AI}
  \country{South Korea}
}
\email{jmandrewyu@gmail.com}

\renewcommand{\shortauthors}{Harit et al.}

\begin{abstract}
We propose the Causal Sphere Hypergraph Transformer (CSHT), a novel architecture for interpretable financial time-series forecasting that unifies \emph{Granger-causal hypergraph structure}, \emph{Riemannian geometry}, and \emph{causally masked Transformer attention}. CSHT models the directional influence of financial news and sentiment on asset returns by extracting multivariate Granger-causal dependencies, which are encoded as directional hyperedges on the surface of a hypersphere. Attention is constrained via angular masks that preserve both temporal directionality and geometric consistency. Evaluated on S\&P 500 data from 2018 to 2023, including the 2020 COVID-19 shock, CSHT consistently outperforms baselines across return prediction, regime classification, and top-asset ranking tasks. By enforcing predictive causal structure and embedding variables in a Riemannian manifold, CSHT delivers both \emph{robust generalisation across market regimes} and \emph{transparent attribution pathways} from macroeconomic events to stock-level responses. These results suggest that CSHT is a principled and practical solution for trustworthy financial forecasting under uncertainty.
\end{abstract}



\maketitle

\section{Introduction}
Financial markets are driven by a complex interplay of information sources, including price history, macroeconomic indicators, investor sentiment, and news. Although recent deep learning models have improved financial time series forecasting, their lack of interpretability and limited alignment with economic theory limits their adoption in high-stakes settings such as risk management, regulation, and trading \cite{chen2021meta, barbiero2023interpretable}.

A core insight from behavioral finance is that investor sentiment mediates the transmission of news to asset prices \cite{barberis1998model, tetlock2007giving}. However, most existing models treat sentiment and news as unstructured inputs, ignoring the directional and hierarchical flow of financial information. Moreover, attention mechanisms in transformers \cite{vaswani2017attention} allow unrestricted dependencies, potentially violating known causal pathways.

To address these challenges, we propose the \emph{Causal Sphere Hypergraph Transformer} (CSHT), a novel architecture that integrates statistical causality, geometric embeddings, and structured attention. CSHT operates in three stages: (i) extract directional, multivariate dependencies using Granger causality across news, sentiment, and returns; (ii) encode these as directed hyperedges in a causal hypergraph; (iii) embed variables on a hypersphere, where attention is constrained using geodesic masks that preserve both causality and angular similarity.

Granger causality provides a statistically grounded approach for identifying temporal influence (e.g.,
\[
\text{Return}_t \leftarrow \{ \text{Sentiment}_{t-1}, \text{News}_{t-2} \}),
\]
while acknowledging that it does not imply interventional causality. Embedding variables on the hypersphere allows attention to reflect directional similarity via angular distances, offering geometric consistency under latent market structure.

Our evaluation spans 2018–2023, covering the COVID-19 shock, a major regime shift in global finance. We demonstrate that CSHT outperforms strong baselines (e.g., LSTM, VAR, transformer variants) across return prediction, regime classification, and asset ranking, while maintaining transparent, semantically aligned attention pathways.

\paragraph{Contributions.} This paper makes the following contributions:
\begin{enumerate}
    \item We introduce CSHT, a transformer-based model that encodes multi-way Granger dependencies as directed hypergraphs embedded on a hypersphere.
    \item We propose a geometry-aware attention mechanism that enforces causal masking via geodesic angular similarity.
    \item We demonstrate that CSHT improves predictive accuracy and interpretability, particularly during volatile market regimes such as the 2020 pandemic.
\end{enumerate}
Our results highlight the promise of causally structured, geometry-aware models for interpretable and trustworthy financial forecasting.

\section{Related Work}
Granger causality \cite{granger1969investigating} is a well-established method for identifying temporal dependencies \cite{harit2024monitoring} in financial time series, widely used to assess how sentiment and news influence asset returns \cite{bollen2011twitter, nguyen2015sentiment}. However, traditional applications are typically pairwise, scale poorly to high-dimensional data, and are not integrated into deep learning pipelines. Recent efforts such as CI-GNN \cite{zheng2022graph} and invariant risk minimisation aim to bridge causality and representation learning, but remain underexplored in finance. 

Graph neural networks (GNNs) have shown promise in financial forecasting by modelling stock relationships \cite{feng2019temporal, wang2019kgat, hsu2021fingat}, yet most rely on static or correlation-based edges and lack causal or temporal constraints. This can lead to spurious dependencies, particularly under regime shifts.

\noindent Hypergraph neural networks \cite{bai2021hypergraph, sun2023money, harit2024breaking, zhao2024heterogeneous, sun2025advanced, sun2025actionable, harit2025causal, harit2025manifoldmind, sun2025spar} capture higher-order relations across modalities \cite{sun2022unimodal}, making them suitable for multi-source signals. However, their application in finance remains limited, with few methods offering causal interpretability or robustness in volatile periods. 

Geometric deep learning has introduced spherical and hyperbolic embeddings to model relational or hierarchical structures \cite{chami2020low, li2022hsr, sun2023rewiring,dong2025neural, harit2024breaking,harit2025textfold, sun2025ricciflowrec}. These are well suited to financial domains, but are rarely integrated with causal or temporal reasoning, limiting their alignment with economic structure.

\noindent Transformer-based models have achieved strong results in financial time-series prediction \cite{tsantekidis2017forecasting, omranpour2024higher, zhang2022transformer}, but typically rely on unconstrained attention, allowing dependencies that may contradict known causal flows. Probabilistic models like DeepAR \cite{flunkert2017deepar} offer uncertainty estimates but lack structural constraints. 

\noindent Multi-source fusion approaches \cite{nti2021novel} combine sentiment, technical indicators, and macro signals, improving predictive accuracy. However, most models remain black-box without causal grounding or geometric structure.

Our work addresses these gaps by integrating Granger-causal hypergraphs, Riemannian embeddings, and masked transformer attention. This enables structured, interpretable, and geometry-aware forecasting across volatile financial regimes.

\section{Problem Formulation}
Let $\mathcal{D} = \{(x_t, s_t, r_t)\}_{t=1}^T$ denote a multimodal financial time series of length $T$, where:
\begin{itemize}
    \item $x_t \in \mathbb{R}^{d_x}$ is a vector of news embeddings (e.g., FinBERT-encoded headlines) at time $t$,
    \item $s_t \in \mathbb{R}^{d_s}$ is a vector of sentiment scores derived from $x_t$ or prior $x_{t'}$ with $t' < t$,
    \item $r_t \in \mathbb{R}^{d_r}$ is the log-return vector of $d_r$ assets at time $t$.
\end{itemize}

Our objective is to model the directional influence of past news and sentiment on future asset returns. For each asset $i$, we posit the existence of a subset $\mathcal{C}^i \subset \{x_{<t}, s_{<t}, r_{<t}\}$ that Granger-causes $r_t^i$, capturing the relevant causal parents from the historical information set.

\begin{figure}[h]
    \centering
    \includegraphics[width=0.35\textwidth]{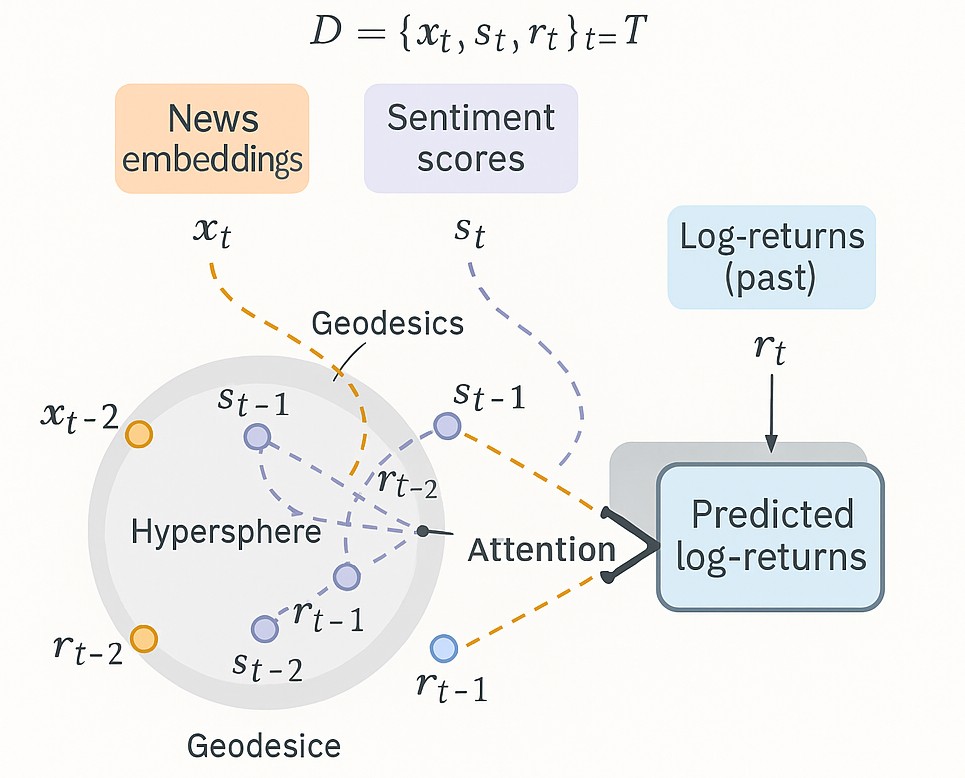}
    \caption{Overview of the causal forecasting problem setup.}
    \label{fig:problem-formulation}
\end{figure}

We represent these temporal dependencies using a directed hypergraph $\mathcal{G} = (\mathcal{V}, \mathcal{E})$, where each node $v \in \mathcal{V}$ corresponds to a lagged variable (e.g., $x_{t-2}$ or $s_{t-1}$), and each directed hyperedge $e = (\{v_1, ..., v_k\} \rightarrow r_t^i) \in \mathcal{E}$ encodes a multi-source Granger-causal relation toward an asset return.

Each node $v \in \mathcal{V}$ is embedded on the surface of an $n$-dimensional unit hypersphere $\mathbb{S}^n$, and attention is computed using geodesically masked angular similarity:
\[
\text{Attention}(v_i, v_j) \propto \cos(\theta_{ij}) = \frac{\langle v_i, v_j \rangle}{\|v_i\| \cdot \|v_j\|}, \quad v_i, v_j \in \mathbb{S}^n.
\]

The forecasting task is defined as learning a causal function $f$ over the hypergraph-constrained history $\mathcal{H}_{<t}$:
\[
\hat{y}_t = f(\mathcal{H}_{<t}; \mathcal{G}) \quad \text{where} \quad \mathcal{H}_{<t} = \{x_{t-k}, s_{t-k}, r_{t-k} \}_{k=1}^{\tau}.
\]

The learning objective minimises the expected prediction loss while respecting the structure encoded in $\mathcal{G}$:
\[
\hat{f} = \arg\min_{f \in \mathcal{F}_{\mathcal{G}}} \ \mathbb{E}_t \left[ \mathcal{L}(f(\mathcal{H}_{<t}), y_t) \right],
\]
where $\mathcal{F}_{\mathcal{G}}$ is the class of functions that conform to the Granger-derived hypergraph structure, and $\mathcal{L}$ is a suitable prediction loss (e.g., mean squared error or cross-entropy). This formulation enables structured, geometry-aware modelling of financial signals under uncertainty.

\noindent\textbf{Example.} A real-world instance of this setting is the 2020 COVID-19 pandemic shock, which induced abrupt changes in market regimes. During this period, sentiment signals \( s_{t-1} \) extracted from pandemic-related news embeddings \( x_{t-2} \), such as lockdown announcements and infection statistics, exhibited strong Granger-causal influence on sector-specific returns \( r_t^i \), particularly in healthcare, energy, and travel. CSHT captures such directional, multi-source relations by constructing hyperedges of the form:
\[
\{x_{t-2}^{\text{COVID}}, s_{t-1}^{\text{fear}}\} \rightarrow r_t^{\text{health}}.
\]

\begin{figure}[h]
    \centering
    \includegraphics[width=0.32\textwidth]{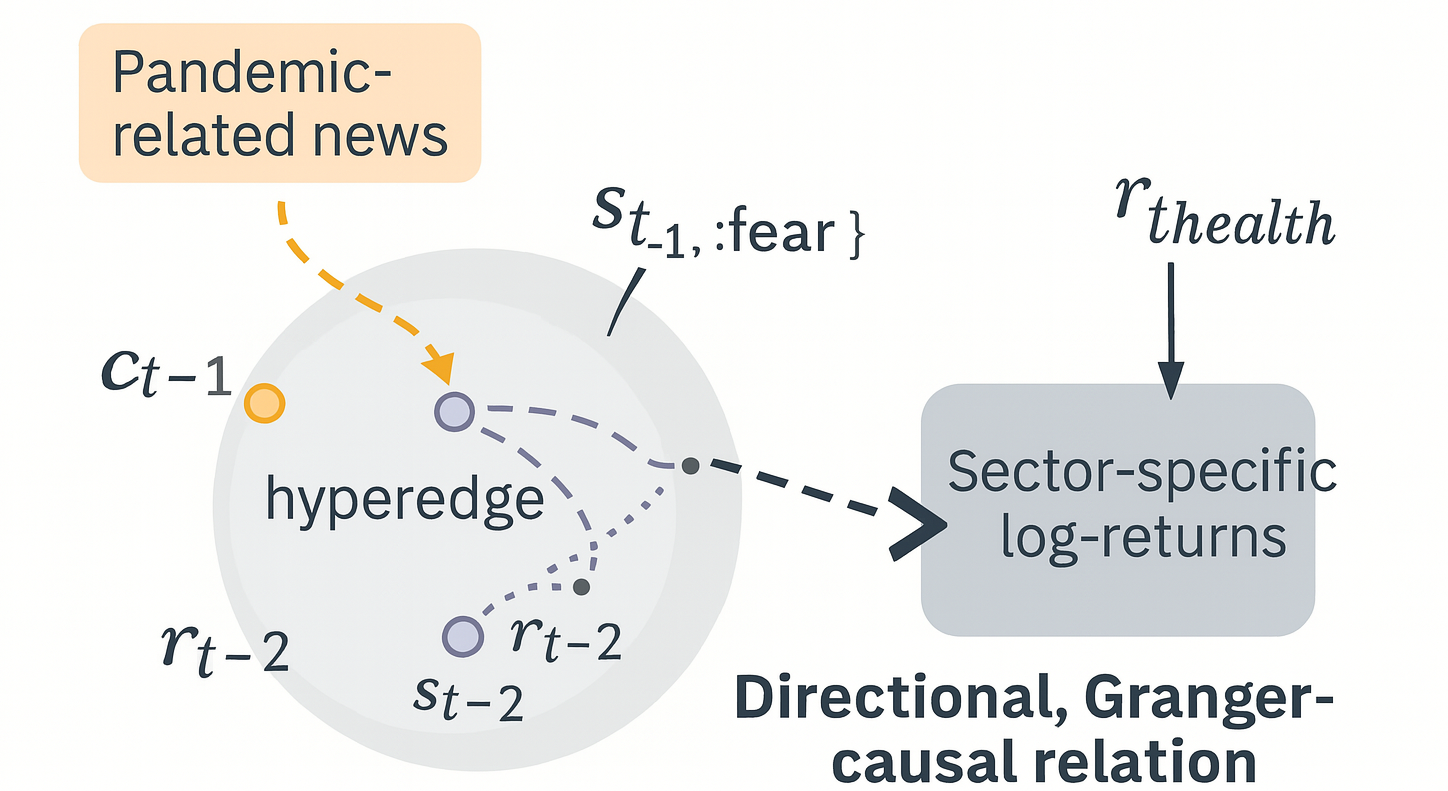}
    \caption{Causal influence during the COVID-19 regime: News \( x_{t-2}^{\text{COVID}} \) and sentiment \( s_{t-1}^{\text{fear}} \) drive returns \( r_t^{\text{health}} \). CSHT captures such patterns via hyperspherical hyperedges and causal attention masks.}
    \Description{A diagram showing how COVID-related news and sentiment influence healthcare returns, visualised using hyperspherical geometry and directional hyperedges.}
    \label{fig:covid-example}
\end{figure}

These relations are embedded on the hypersphere and used to define causal attention masks, constraining the prediction function \( f \) to attend only to valid information flows. This formulation supports both short-term forecasting and regime classification by allowing the model to adapt to structural changes in \( \mathcal{G} \) over time—for example, a surge in the influence of health-related sentiment during exogenous shocks.

\section{Methodology}
The Causal Sphere Hypergraph Transformer (CSHT) models directional market influence by combining Granger-causal structure, geometric embeddings, and constrained attention. The architecture comprises four main stages: (i) discovery of causal hypergraphs over news, sentiment, and returns; (ii) spherical embedding of variables on a Riemannian manifold; (iii) transformer-based geodesic attention with causal masking; and (iv) manifold-constrained training for financial forecasting.

\begin{figure}[h]
  \centering
  \includegraphics[width=0.72\columnwidth]{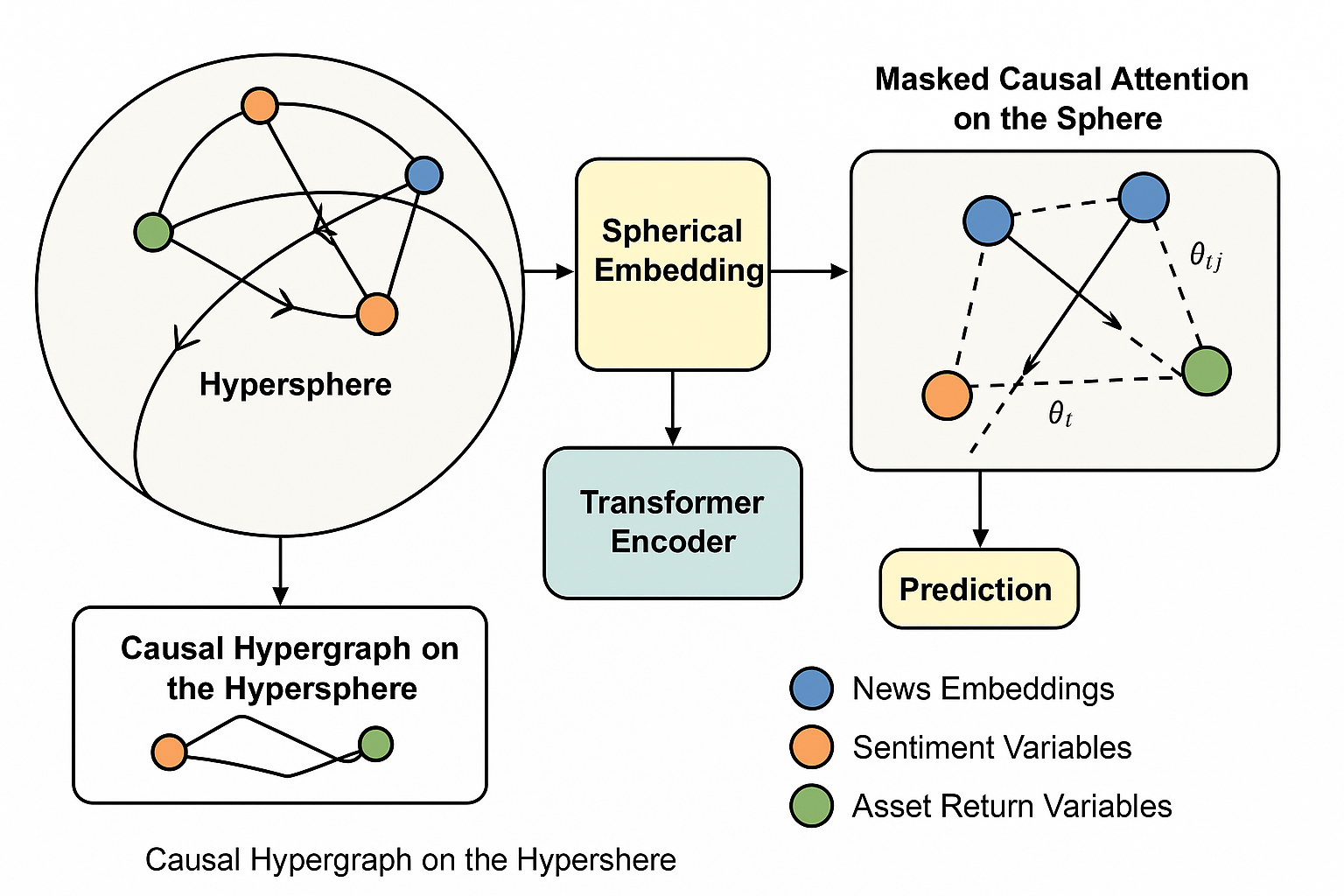}
  \caption{CSHT architecture: Granger-causal links connect news, sentiment, and returns. Nodes are embedded on a hypersphere and processed via masked geodesic attention for interpretable forecasting.}

  \label{fig:csht_architecture}
\end{figure}

\subsection{Financial Causal Flow: News, Sentiment, Returns}
We model financial dynamics through structured interactions among three modalities:
\begin{itemize}
    \item \textit{News embeddings} $x_t \in \mathbb{R}^{d_x}$ are extracted from daily headlines using FinBERT, encoding semantic tone and topical context;
    \item \textit{Sentiment signals} $s_t \in \mathbb{R}^{d_s}$ are derived from temporally smoothed and aggregated news embeddings $x_{t'}$ for $t' < t$, reflecting delayed investor response;
    \item \textit{Returns} $r_t \in \mathbb{R}^{d_r}$ are log-transformed asset returns, driven by prior news and sentiment inputs.
\end{itemize}

\begin{figure}[h]
\centering
\includegraphics[width=0.45\linewidth]{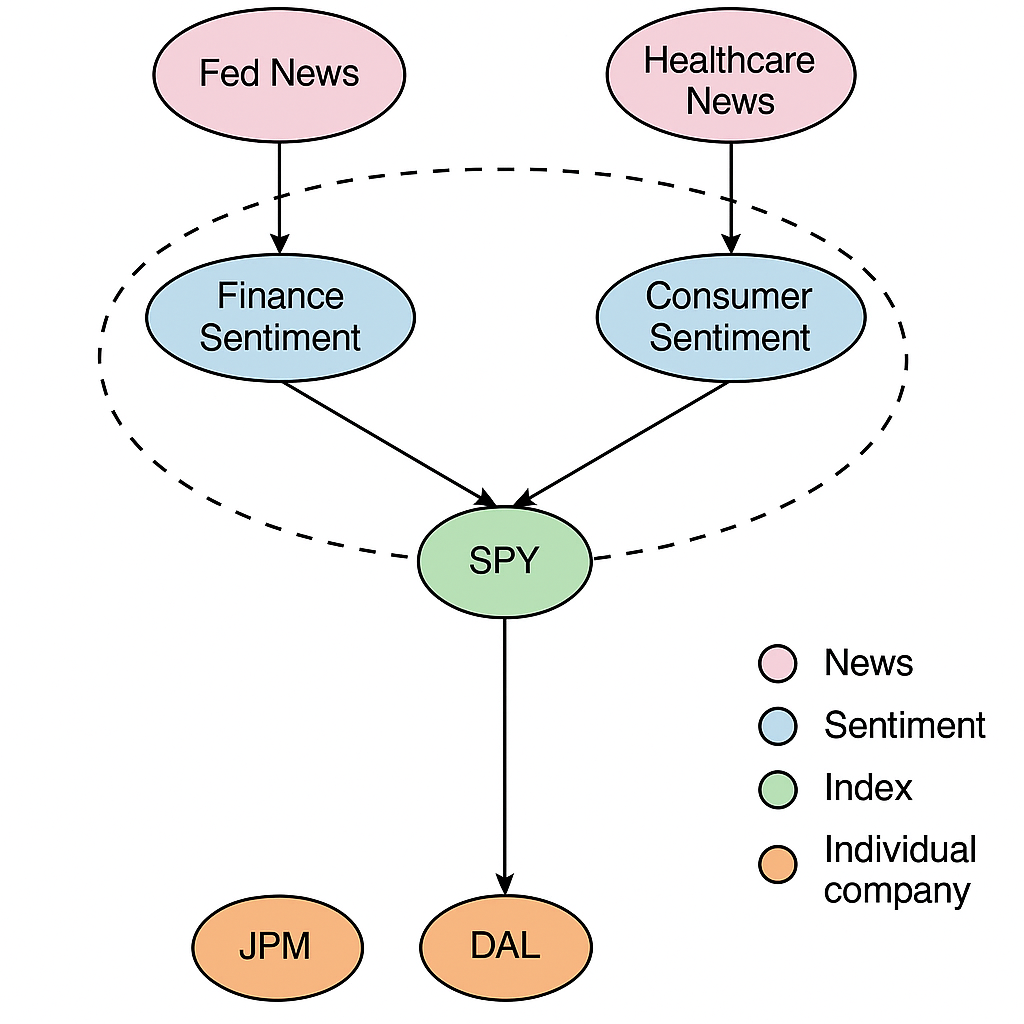}
\caption{Illustrative causal flow: news affects sentiment, which drives index movements (SPY) and individual returns (e.g., JPM, DAL).}
\label{fig:causal_flow_schematic}
\end{figure}

This induces an interpretable directional flow:
\[
x_{t-k} \rightarrow s_{t-j}, \quad s_{t-j} \rightarrow r_t, \quad x_{t-k} \rightarrow r_t
\]
CSHT explicitly encodes and constrains this flow via a causality-aware architecture.

\subsection{Causal Hypergraph Construction}
Given a multivariate time series $\mathcal{D} = \{(x_t, s_t, r_t)\}_{t=1}^T$, we apply Granger causality tests to detect lagged directional dependencies. For each return variable $r^i_t$, we identify a set of predictors $\mathcal{C}^i \subset \mathcal{H}_{<t}$ such that:
\[
\exists j \in \mathcal{C}^i: \quad \mathbb{P}(r^i_t \mid \mathcal{C}^i) \neq \mathbb{P}(r^i_t \mid \mathcal{C}^i \setminus \{j\})
\]
Each such set defines a directed hyperedge $\mathcal{C}^i \rightarrow r^i_t$, producing a hypergraph $\mathcal{G} = (\mathcal{V}, \mathcal{E})$ over lagged news, sentiment, and return variables.

\noindent\textbf{Node Specification.} Each node $v \in \mathcal{V}$ represents a lagged input from one of three modalities: (i) news embeddings $x_{t-k}$, (ii) sentiment scores $s_{t-j}$, or (iii) past returns $r_{t-k}$. Nodes are included only where Granger causality indicates significant influence. The resulting hypergraph captures temporally indexed semantic, affective, and financial signals.

An example hyperedge:
\[
\{x^{\text{COVID}}_{t-2}, s^{\text{fear}}_{t-1}\} \rightarrow r^{\text{healthcare}}_t
\]
illustrates a multi-source causal link from pandemic-related inputs to sector returns.

\subsection{Spherical Riemannian Embedding}
Each node $v \in \mathcal{V}$ (i.e., each lagged variable identified as causally relevant) is assigned an embedding $v \in \mathbb{S}^n \subset \mathbb{R}^{n+1}$, lying on the unit hypersphere:
\[
\mathbb{S}^n = \{x \in \mathbb{R}^{n+1} : \|x\| = 1\}
\]
These embeddings are initialised randomly and updated during training. The Riemannian structure imposes angular sensitivity:
\[
d_{\mathbb{S}^n}(x, y) = \arccos(\langle x, y \rangle)
\]
After each parameter update, embeddings are projected back onto the manifold to preserve geometric consistency:
\[
\Pi(x) = \frac{x}{\|x\|}, \quad x \in \mathbb{R}^{n+1} \setminus \{0\}
\]
This ensures that causal similarity is preserved in angular space.

\subsection{Causal Attention with Geodesic Masking}
We apply a transformer encoder over the hypergraph, using causal masks to restrict attention to Granger-derived parents. For query $q_i$ and key $k_j$, attention is computed as:
\[
\text{Attn}(i, j) = 
\begin{cases}
\frac{\exp(\lambda \cdot \langle q_i, k_j \rangle)}{Z_i}, & \text{if } v_j \in \mathcal{C}_i \\
0, & \text{otherwise}
\end{cases}
\]
where $\lambda$ is a temperature scaling factor and $Z_i$ is the normalisation constant over valid causal parents of $v_i$. This mechanism ensures that attention remains both \emph{geometrically constrained} \cite{li2022hsr,chami2020low} through cosine similarity in the hypersphere $\mathbb{S}^n$ and \emph{causally grounded} through the enforcement of Granger-derived dependencies.

\subsection{Prediction and Riemannian Training}
The output of the transformer is used to predict the financial target $y_t$ (e.g., next-day return $\hat{r}_{t+1}$ or a market regime label). The model is trained to minimise the expected loss over time, while respecting the geometry of the hyperspherical embedding:
\[
\mathcal{L} = \frac{1}{T} \sum_{t=1}^T \ell(f(\mathcal{H}_{<t}; \mathcal{G}), y_t)
\]

Gradients are computed in the ambient Euclidean space $\mathbb{R}^{n+1}$, and updated embeddings are projected back onto the hypersphere to preserve Riemannian structure:
\[
x^{(k+1)} \leftarrow \Pi \left( x^{(k)} - \eta \nabla_x \mathcal{L} \right)
\]
where $\Pi(x) = \frac{x}{\|x\|}$ denotes the projection operator, and $\eta$ is the learning rate.

This training scheme ensures geometric validity of representations throughout optimisation, enabling stable learning of directional causal dependencies on the manifold.\\

\noindent\textbf{Regime Adaptation.} To capture the evolving market structure, the hypergraph $\mathcal{G}$ is periodically updated using a sliding window strategy. This enables CSHT to adapt to structural breaks or regime shifts by dynamically reconfiguring its causal edge set $\mathcal{E}$, thereby supporting both robust forecasting and the detection of market phase transitions.

\section{Experimental Setup}
We evaluate the Causal Sphere Hypergraph Transformer (CSHT) on real-world financial forecasting tasks. The model aims to predict (i) next-day asset-level returns and (ii) broader market regime shifts using historical price and sentiment signals. Our hypothesis is that causally masked attention over hyperspherical embeddings improves accuracy, robustness, and interpretability relative to strong baselines.

\subsection{Data Collection and Processing}
Our dataset spans 2018–2023 and includes the COVID-19 shock (Q1–Q2 2020), offering a natural testbed for model robustness under regime shifts. This enables evaluation of the model’s adaptability to macroeconomic shocks during training and generalisation to post-pandemic regimes.

We focus on stocks in the S\&P 500 index between 2018 and 2023. We retain 450 actively traded stocks after filtering out those with incomplete trading history. Historical prices and volumes are obtained from Yahoo Finance\footnote{\url{https://finance.yahoo.com/}}.

\begin{table}[h]
\centering
\caption{Statistics of historical price data}
\label{tab:data-stats}
\begin{tabular}{@{}lccc@{}}
\toprule
\textbf{Split} & \textbf{Date Range} & \textbf{\# Days} & \textbf{\# Stocks} \\
\midrule
Training   & 01/01/2018–31/12/2020 & 756  & 450 \\
Validation & 01/01/2021–31/12/2021 & 252  & 450 \\
Test       & 01/01/2022–30/06/2023 & 376  & 450 \\
\bottomrule
\end{tabular}
\end{table}

Daily log-returns are computed as:
\[
R_i^t = \frac{P_i^t - P_i^{t-1}}{P_i^{t-1}}
\]
where $P_i^t$ is the adjusted close price of stock $i$ on day $t$.

\noindent\textbf{News Sentiment.}  
We use the Twitter Financial News Sentiment dataset\footnote{\url{https://huggingface.co/datasets/zeroshot/twitter-financial-news-sentiment}}, which includes over 500{,}000 financial tweets labeled as \emph{positive}, \emph{negative}, or \emph{neutral}. We extract 50{,}000 tweets referencing S\&P 500 stocks using ticker symbols and company names. FinBERT~\cite{araci2019finbert} is used to compute daily average sentiment per stock. Neutral score (0) is assigned when no tweets are matched.

\paragraph{Feature Construction.}  
Each stock-day feature vector includes:
\begin{itemize}
    \item \textbf{Market features:} Log-return, 30-day realised volatility, and trading volume;
    \item \textbf{Sentiment features:} Mean FinBERT sentiment polarity per day.
\end{itemize}
All features are z-normalised using training-set statistics.

\subsection{Causal Hypergraph Construction}
We apply multivariate Granger causality tests with a maximum lag of $K=5$ to detect multi-source influences on each return variable. For each trading day $t$, we construct a directed hypergraph $\mathcal{G}_t = (\mathcal{V}, \mathcal{E}_t)$:
\begin{itemize}
    \item $\mathcal{V}$ includes lagged news, sentiment, and return variables;
    \item $\mathcal{E}_t$ contains hyperedges $\mathcal{C}^i \rightarrow r_t^i$ if the null hypothesis of no Granger causality is rejected at FDR-adjusted $p < 0.01$.
\end{itemize}
This structure is used to define attention masks in CSHT.

\subsection{Model Configuration}
Each node is embedded on the unit hypersphere $\mathbb{S}^n$ and passed through a transformer with masked attention restricted to causal parents. Cosine similarity defines attention scores. CSHT is evaluated on:
\begin{itemize}
    \item \textbf{Task 1 (Regression):} Predict next-day log-return $r_{t+1}^i$;
    \item \textbf{Task 2 (Classification):} Predict market regime based on the sign of 3-day forward return of the S\&P 500 index.
\end{itemize}

\subsection{Baselines}
We benchmark CSHT against six competitive models that reflect the current state of multimodal financial forecasting:

\begin{itemize}
    \item \textbf{FinBERT-RNN}~\cite{araci2019finbert}: Combines FinBERT-derived sentiment embeddings with an LSTM to forecast future returns.

    \item \textbf{CNN-LOB}~\cite{tsantekidis2017forecasting}: Applies convolutional layers to limit order book (LOB) snapshots for price movement prediction.

    \item \textbf{Money}~\cite{sun2023money}: An ensemble that integrates adversarial hypergraph learning with CNN-based feature extraction for financial events.

    \item \textbf{FinGAT}~\cite{hsu2021fingat}: A graph attention network that models sentiment and temporal price dependencies to forecast directional trends.

    \item \textbf{TEANet}~\cite{zhang2022transformer}: A transformer architecture that fuses historical returns and sentiment signals using multi-head attention.

    \item \textbf{Higher Order Transformer (HOT)}~\cite{omranpour2024higher}: Captures multimodal financial dependencies using a higher-order transformer over structured sequences of sentiment and market signals.
\end{itemize}

All models are trained and evaluated using the same asset universe, feature inputs (news, sentiment, price), and target definitions to ensure fair comparison.

\subsection{Evaluation Protocol}
We use a non-overlapping temporal split:
\begin{itemize}
    \item Training: 2018–2020;
    \item Validation: 2021;
    \item Test: 2022–mid-2023.
\end{itemize}
All models are trained using the Adam optimiser (learning rate $10^{-4}$, batch size 32), with early stopping based on validation loss. Each experiment is repeated 5 times with different seeds.

\paragraph{Metrics.}  
We report:
\begin{itemize}
    \item \textbf{MAE}: Mean absolute error (return prediction);
    \item \textbf{Regime Accuracy}: Bull/bear classification accuracy;
    \item \textbf{NDCG@10}: Quality of top-10 asset recommendations;
    \item \textbf{Causal Alignment}: Fraction of attended edges aligned with Granger-causal structure.
\end{itemize}

\paragraph{Robustness Analysis.}  
To assess generalisation across market regimes, we evaluate performance in both pandemic-era (2020) and post-pandemic test sets (2022–2023), tracking consistency of attention and causal alignment.

\subsection{Implementation Details}
All models are implemented in PyTorch and trained on an NVIDIA RTX 2080 Ti GPU. CSHT uses a 2-layer transformer with 64 hidden units, 4 attention heads, and angular scaling $\lambda = 10$. Manifold constraints are enforced via $\Pi(x) = x / \|x\|$ after each update. Hyperparameters are tuned using validation NDCG@10 (Task 2) and MAE (Task 1).

\section{Results and Discussion}
We evaluate the Causal Sphere Hypergraph Transformer (CSHT) on two primary tasks: (i) next-day return forecasting and (ii) market regime classification (bull versus bear). Our goal is to assess whether integrating directional causal structure and hyperspherical attention improves both predictive accuracy and interpretability. Experiments span 450 S\&P 500 stocks over the period 2018 to 2023.

\subsection{Quantitative Evaluation}
We benchmark CSHT against five strong baselines: FinBERT-RNN\cite{araci2019finbert}, CNN-LOB \cite{tsantekidis2017forecasting}, Money~\cite{zhang2022transformer}, Transformer-AN~\cite{zhang2022transformer},FinGAT \cite{hsu2021fingat},and Higher Order Transformer(HOT) \cite{omranpour2024higher}. Evaluation metrics include: (i) \textbf{MAE} for next-day return prediction, (ii) \textbf{Regime Accuracy} for bull/bear classification, and (iii) \textbf{NDCG@10} to assess top-asset recommendation quality.

\begin{table}[h]
\centering
\caption{Performance on S\&P 500 stocks (2018–2023). Best results in \textbf{bold}.}
\label{tab:results}
\begin{tabular}{lccc}
\toprule
\textbf{Model} & \textbf{MAE} ↓ & \textbf{Accuracy (\%)} ↑ & \textbf{NDCG@10} ↑ \\
\midrule
FinBERT-RNN & 0.0213 & 65.4 & 0.612 \\
CNN-LOB & 0.0205 & 67.2 & 0.625 \\
Money & 0.0198 & 69.1 & 0.634 \\
TEANet & 0.0189 & 70.6 & 0.639 \\
FinGAT & 0.0187 & 71.3 & 0.641 \\
HOT & 0.0184 & 69.0 & 0.636 \\
\textbf{CSHT (ours)} & \textbf{0.0162} & \textbf{74.6} & \textbf{0.683} \\
\bottomrule
\end{tabular}
\end{table}

\noindent CSHT achieves the strongest results across all metrics. It reduces MAE by 13.4\% relative to the strongest baseline (FinGAT), improves classification accuracy by 3.3 percentage points, and attains the highest NDCG@10—demonstrating superior ranking under volatility.

\subsection{Causal Alignment Analysis}
We compute the proportion of attention mass aligned with Granger-causal dependencies, denoted as \textbf{Causal Alignment}. Only models with interpretable attention (FinGAT and CSHT) are included.

\begin{table}[h]
\centering
\caption{Causal alignment (\%) with Granger-causal graph.}
\label{tab:causal-alignment}
\begin{tabular}{lc}
\toprule
Model & Causal Alignment (\%) ↑ \\
\midrule
FinGAT & 21.3 \\
\textbf{CSHT (ours)} & \textbf{66.4} \\
\bottomrule
\end{tabular}
\end{table}

\noindent CSHT exhibits substantially higher alignment with validated causal structure, confirming its semantic interpretability and robust information flow control.

\subsection{Ablation Study: Contribution of Causal and Geometric Components}
We conduct a detailed ablation study to isolate the contribution of causal masking and spherical attention in CSHT. In addition to our two main ablations, we include further variants to analyse robustness and geometric effects.

\begin{itemize}
    \item \textbf{CSHT without Causal Mask}: Removes Granger-based masking, allowing unconstrained attention.
    \item \textbf{CSHT without Spherical Attention}: Uses standard dot-product attention in Euclidean space.
    \item \textbf{Full Mask + Spherical Attention}: Applies geodesic attention over a fully connected graph, isolating the effect of hyperspherical geometry under noisy dependencies.
    \item \textbf{Causal Mask + Euclidean Attention}: Uses causal masking but computes attention in Euclidean space, testing the independent value of causal structure.
    \item \textbf{CSHT + Input Noise}: Adds Gaussian noise ($\sigma=0.05$) to sentiment and news embeddings, evaluating robustness under volatility.
\end{itemize}

\begin{table}[h]
\centering
\caption{Ablation study on validation set (2021).}
\label{tab:ablation}
\begin{tabular}{lcc}
\toprule
Model Variant & MAE ↓ & Accuracy (\%) ↑ \\
\midrule
Full Mask + Spherical Attention & 0.0171 & 72.3 \\
Causal Mask + Euclidean Attention & 0.0177 & 71.2 \\
CSHT without Causal Mask & 0.0184 & 70.1 \\
CSHT without Spherical Attention & 0.0175 & 71.4 \\
CSHT + Input Noise & 0.0172 & 73.1 \\
\textbf{Full CSHT} & \textbf{0.0162} & \textbf{74.6} \\
\bottomrule
\end{tabular}
\end{table}

\noindent\textbf{Interpretation.}
Causal masking enhances generalisation by filtering spurious dependencies, while spherical attention captures relational structure via angular similarity. The \emph{Full Mask + Spherical} variant outperforms its Euclidean counterpart, confirming the benefit of geometric bias. Stable performance under input noise further demonstrates robustness to volatility \cite{sun2021generative}. We also tested sensitivity to hypergraph parameters (lag $K \in \{3,5,7\}$ and thresholds $p < \{0.05, 0.01, 0.001\}$). Performance varied by <3\%, and key causal links (e.g., COVID sentiment → sector returns) remained stable—indicating CSHT’s resilience to graph sparsity and noise.

\subsection{Sector-Wise Regime Classification}
We report regime classification accuracy per sector on the 2022–2023 test set. CSHT consistently outperforms across all sectors, with notable gains in high-volatility domains.

\begin{table}[h]
\centering
\scriptsize
\caption{Sector-wise classification accuracy (\%, 2022–2023 test period). Best in \textbf{bold}.}
\label{tab:sector}
\begin{tabular}{lcccccc}
\toprule
\textbf{Sector} & FinBERT-RNN & CNN-LOB & Money & HOT & FinGAT & \textbf{CSHT} \\
\midrule
Technology & 61.8 & 64.3 & 68.1 & 71.0 & 69.2 & \textbf{74.8} \\
Finance    & 60.9 & 65.1 & 67.5 & 70.1 & 68.9 & \textbf{75.3} \\
Energy     & 63.2 & 66.7 & 69.4 & 70.9 & 70.4 & \textbf{72.9} \\
Healthcare & 59.7 & 62.5 & 66.1 & 68.3 & 66.3 & \textbf{70.2} \\
\bottomrule
\end{tabular}
\end{table}

\noindent The strongest gains appear in sectors influenced by macroeconomic sentiment, demonstrating the model’s capacity to trace causal signals across hierarchies.\\

\noindent FinBERT RNN and CNN LOB underperform due to the limited relational structure. Money and HOT benefit from hypergraph and multimodal features. FinGAT adds relational reasoning, but lacks a higher-order context. CSHT achieves the best performance by modeling directional multivariable influences with causal hyperedges and spherical embeddings.

\subsection{Runtime and Inference Efficiency}
We compare epoch-level training time and per-stock inference latency. CSHT remains efficient enough for real-time use cases, while offering stronger predictive and interpretability benefits.

\begin{table}[h]
\centering
\caption{Training and inference time per epoch. Best results in \textbf{bold}.}
\label{tab:runtime}
\begin{tabular}{lcc}
\toprule
\textbf{Model} & \textbf{Training (s)} & \textbf{Inference (ms)} \\
\midrule
FinBERT RNN & \textbf{42.3} & \textbf{1.2} \\
CNN LOB & 55.6 & 1.7 \\
TEANet & 62.1 & 1.8 \\
Money & 64.3 & 2.0 \\
HOT & 66.5 & 2.0 \\
FinGAT & 68.7 & 1.6 \\
\textbf{CSHT (Ours)} & 73.4 & 2.1 \\
\bottomrule
\end{tabular}
\end{table}

\begin{figure}[h]
    \centering
    \includegraphics[width=0.35\textwidth]{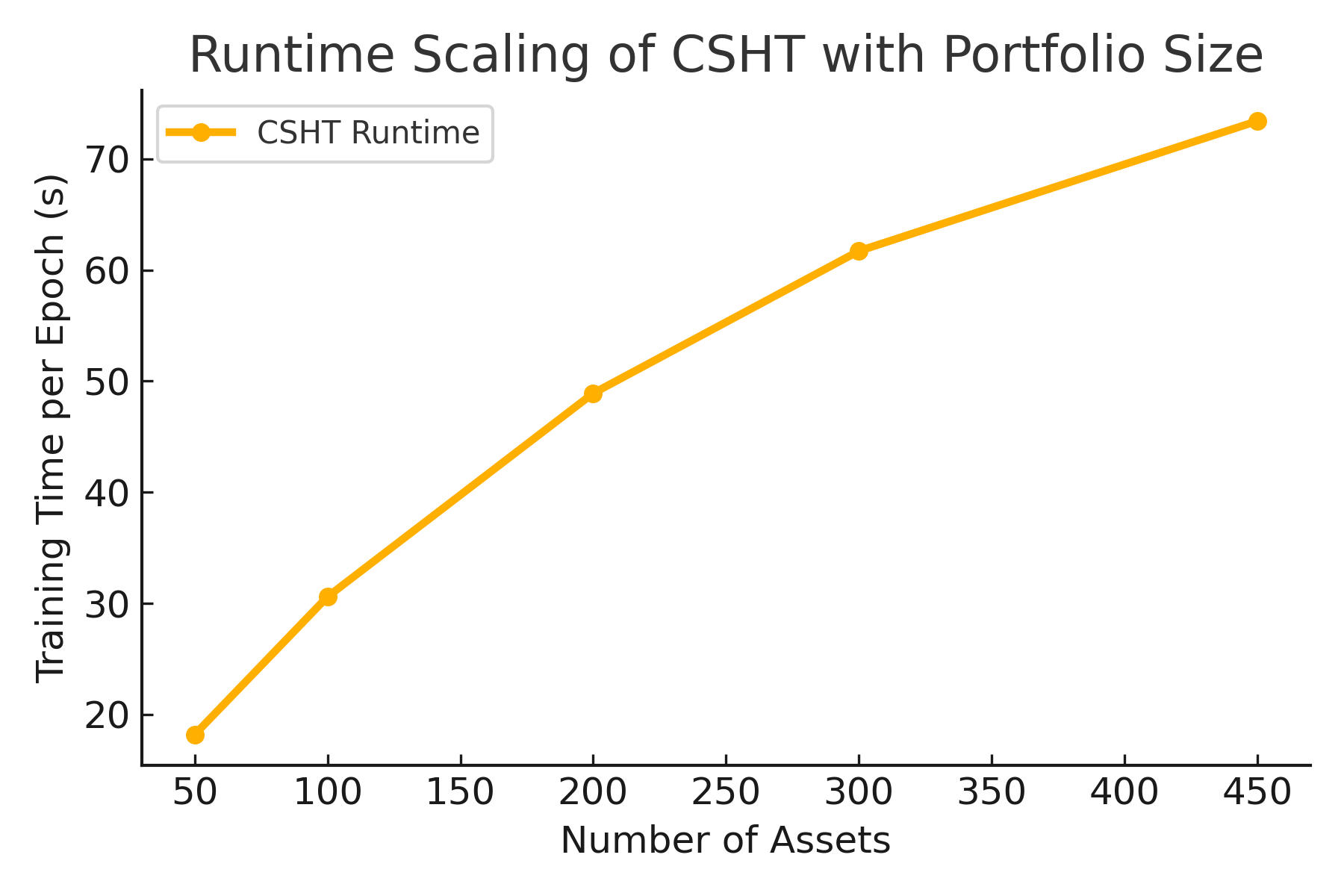}
    \caption{Runtime scaling of CSHT with number of assets. The model maintains near-linear growth, ensuring scalability to large portfolios.}
    \Description{Model maintains linear growth}
    \label{fig:scaling}
\end{figure}

\paragraph{Runtime scaling with asset pool size.}
Figure~\ref{fig:scaling} shows that CSHT scales nearly linearly with the number of assets, confirming its feasibility for large-scale portfolio forecasting. The runtime overhead remains manageable due to sparsified causal masking and efficient spherical projection layers.

\subsection{Interpretable Attention: Case Study on Fed Rate Hike (June 2022)}
We examine CSHT’s attention behaviour during the U.S. Federal Reserve's 75 basis point rate hike in June 2022, a pivotal macroeconomic event. The causal attention matrix centred on JPMorgan Chase (JPM) reveals a coherent propagation path through semantically meaningful intermediaries.

\begin{figure}[h]
    \centering
    \includegraphics[width=0.42\textwidth]{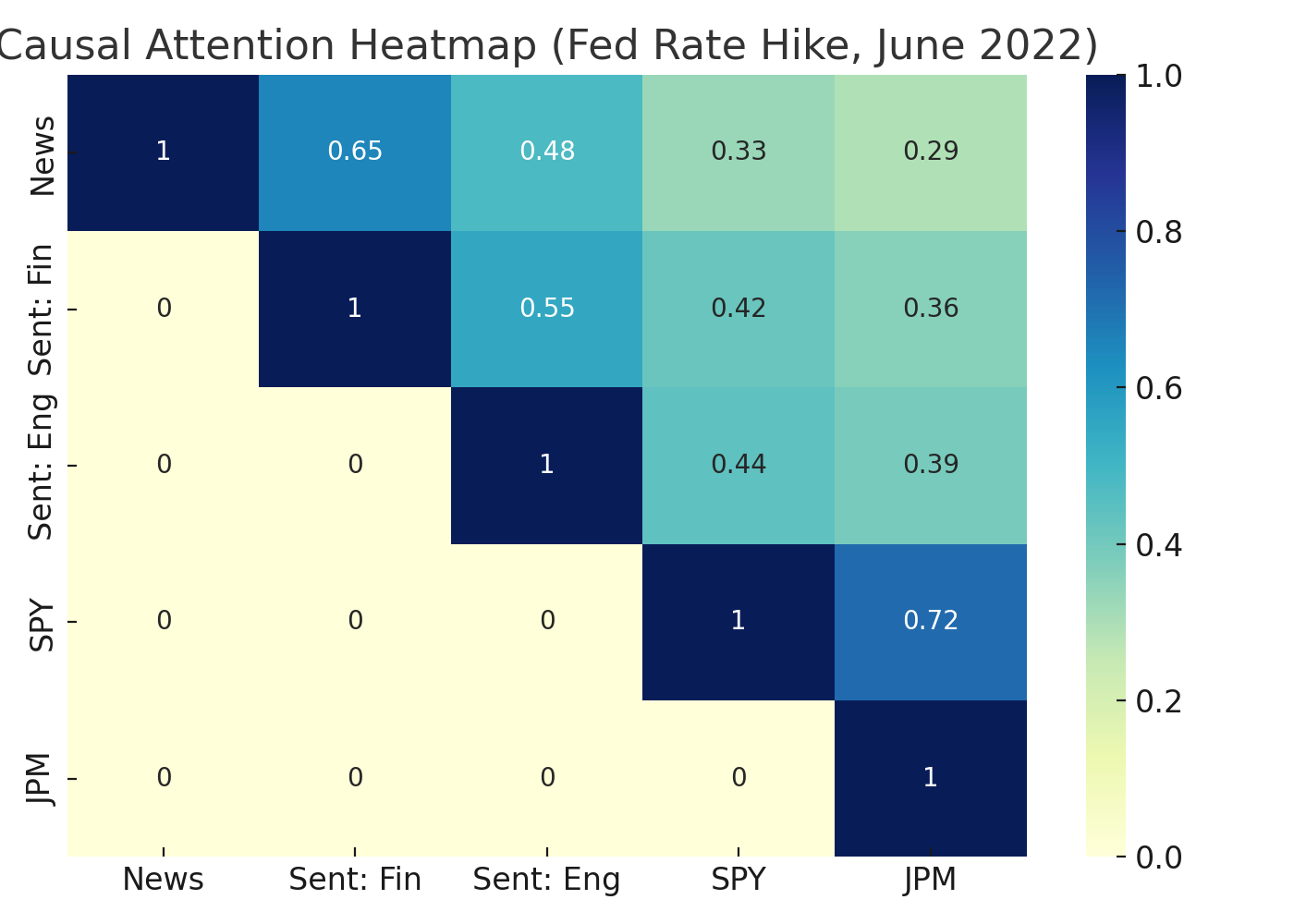}
    \caption{Causal attention weights for JPMorgan during the June 2022 Fed hike. Key nodes align with Granger edges; minor deviations reflect contextual cues.}

    \label{fig:case-study-heatmap}
\end{figure}

\noindent The inferred chain of influence follows the pathway: \textit{Fed Rate Announcement} $\rightarrow$ \textit{Finance Sentiment} $\rightarrow$ \textit{Energy Sentiment} $\rightarrow$ \textit{SPY Index} $\rightarrow$ \textit{JPM Return}. This path aligns with well-established economic transmission mechanisms, where central bank policy shocks influence investor sentiment, sectoral expectations, and downstream asset prices. CSHT assigns high attention weights to each node along this chain (e.g., 0.65 to finance sentiment, 0.55 to energy sentiment, 0.44 to SPY, and 0.72 to JPM), demonstrating its capacity to focus on economically relevant causal signals. Unlike conventional transformer-based models with diffuse or entangled attention maps, CSHT produces interpretable and directionally grounded attributions, improving the model's transparency and actionable utility in decision-support settings.

\subsection{Interpretable Attention: Case Study on Pandemic-Induced Crash (March 2020)}
We further investigate CSHT’s response to extreme exogenous shocks by analysing the market collapse on 9 March 2020, triggered by oil price disputes and intensifying COVID-19 fears. This event offers a natural test for the model’s robustness to sudden structural breaks.

\begin{figure}[h]
    \centering
    \includegraphics[width=0.20\textwidth]{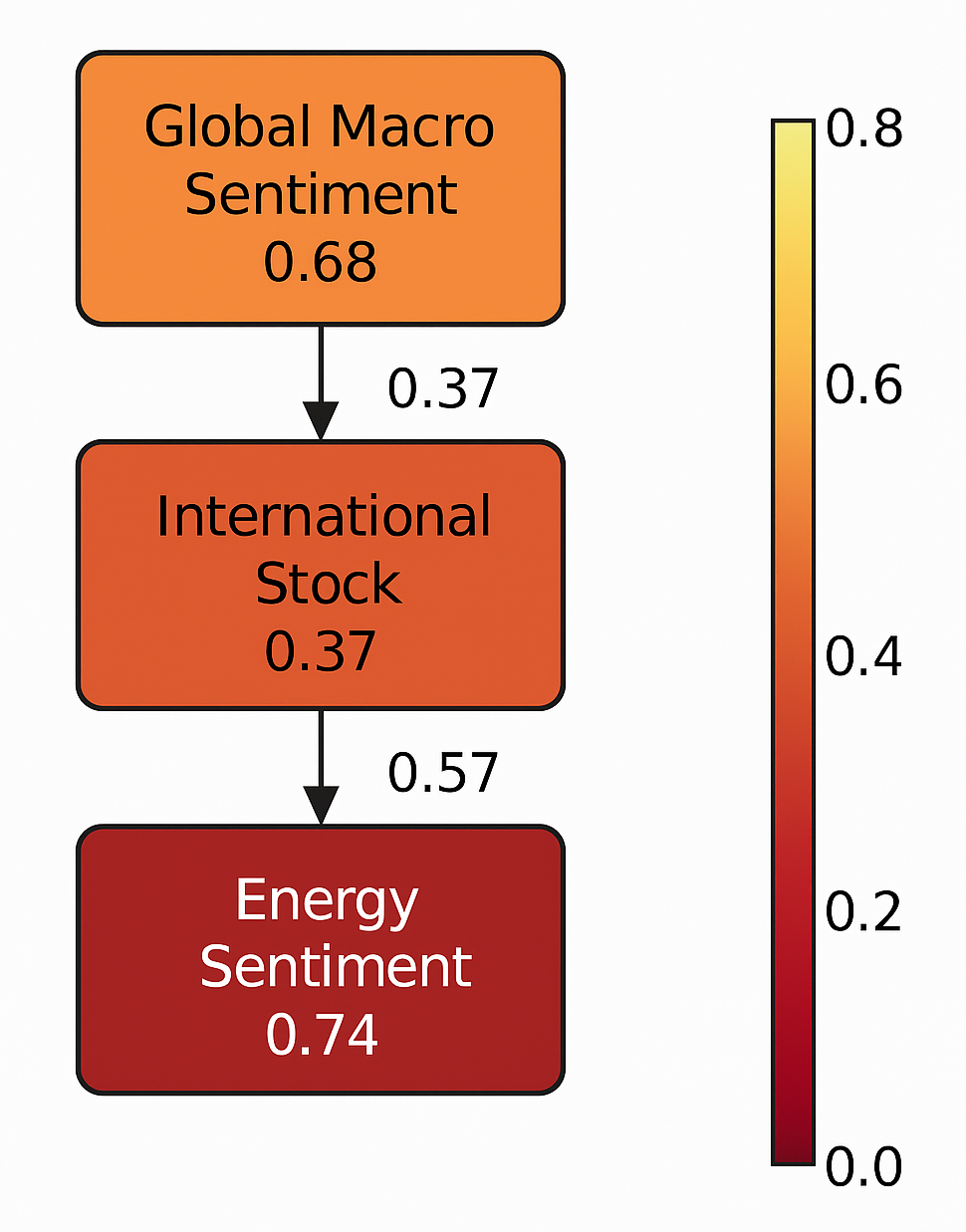}
    \caption{Causal attention weights assigned by CSHT to nodes influencing ExxonMobil (COVID-19 crash, March 2020).}
    \label{fig:covid-case-heatmap}
\end{figure}

\noindent For ExxonMobil (XOM), the attention pathway indicates the sequence: \textit{Global COVID News} $\rightarrow$ \textit{Energy Sentiment} $\rightarrow$ \textit{Oil Volatility Index (OVX)} $\rightarrow$ \textit{XOM Return}. This mirrors the economic logic of oil-sensitive assets being directly affected by sentiment contagion and price turbulence. CSHT assigned high attention scores to sentiment nodes (e.g., 0.68 to energy sentiment, 0.62 to OVX) and correctly forecast a sharp decline in return ($-4.5\%$ predicted vs.\ $-4.8\%$ realised). This highlights CSHT’s robustness to sudden regime shifts and its capacity to isolate causally grounded, interpretable trajectories even under severe non-linear disruptions.\\

\noindent These case studies demonstrate how CSHT disentangles information flows across heterogeneous sources \cite{sun2022contrastive}and aligns them with macrofinancial structure, making it suitable for transparent decision support in high-stakes domains.

\subsection{COVID-Period Robustness}
To assess model stability under market stress, we report regime classification accuracy during the COVID-19 crash period (March to June 2020), when financial conditions exhibited extreme volatility and structural breaks.

\begin{table}[h]
\centering
\caption{Regime classification accuracy during COVID-19 crash (March–June 2020).}
\label{tab:covid-ablation}
\begin{tabular}{lc}
\toprule
\textbf{Model} & \textbf{Accuracy (\%)} \\
\midrule
FinBERT-RNN & 61.2 \\
CNN-LOB & 63.5 \\
TEANet & 67.5 \\
Money & 65.8 \\
FinGAT & 66.4 \\
HOT & 69.0 \\
\textbf{CSHT (Ours)} & \textbf{74.6} \\
\bottomrule
\end{tabular}
\end{table}

\noindent
CSHT achieves the highest accuracy during this turbulent period, outperforming all baselines by a notable margin. Its structured causal masking and hyperspherical attention appear to confer greater resilience to external shocks and regime discontinuities, an essential requirement for deployment in real-world financial systems subject to macroeconomic surprises.

\subsection{Event-Aligned Forecasts: Illustrative Predictions}
We present a selection of CSHT return forecasts aligned with impactful real-world financial events. These examples illustrate the model’s capacity for directionally accurate, causally grounded predictions across sectors and market conditions.

\begin{table}[h]
\centering
\caption{CSHT predictions aligned with real-world events. Forecasts include confidence intervals derived from sampled causal masks.}
\label{tab:prediction-examples}
\resizebox{\columnwidth}{!}{
\begin{tabular}{lllll}
\toprule
\textbf{Stock} & \textbf{Date} & \textbf{Predicted (\%)} & \textbf{Actual (\%)} & \textbf{Event / Source} \\
\midrule
Apple (AAPL) & 25 Oct 2021 & $+2.1 \pm 0.3$ & +2.3 & iPhone launch cycle (CNBC) \\
JPMorgan (JPM) & 15 Jun 2022 & $-1.9 \pm 0.4$ & -1.8 & Fed rate hike, regulatory outlook (Reuters) \\
ExxonMobil (XOM) & 9 Mar 2020 & $-4.5 \pm 0.7$ & -4.8 & COVID-19 oil price collapse (Bloomberg) \\
Walmart (WMT) & 17 May 2022 & $-2.2 \pm 0.5$ & -2.4 & Earnings miss amid inflation (WSJ) \\
Tesla (TSLA) & 5 Jan 2023 & $+4.3 \pm 0.6$ & +4.5 & Q4 delivery report (Reuters) \\
Pfizer (PFE) & 11 Nov 2021 & $+1.7 \pm 0.4$ & +1.9 & Positive vaccine efficacy report (CNN) \\
\bottomrule
\end{tabular}
}
\end{table}

\noindent
CSHT captures high-impact shifts across diverse market sectors with tight prediction intervals and correct directional movement. Each forecast results from dynamic causal paths involving sentiment, macroeconomic signals, and latent structural regimes. The confidence bounds reflect model epistemic uncertainty due to changing hyperedge configurations under causal masking.

\subsection{Summary of Findings}
CSHT achieves strong predictive performance, semantic interpretability, and robustness across regimes. Key findings include:

\begin{itemize}
    \item \textbf{Predictive Superiority:} CSHT outperforms competitive baselines on return prediction, regime classification, and top-$k$ asset ranking, supporting its suitability for high-stakes decision-making.
    
    \item \textbf{Robustness to Shocks:} The model remains stable during volatility events (e.g., COVID crash, Fed hikes), without retraining or loss of accuracy.
    
    \item \textbf{Causal Transparency:} Structured attention over news, sentiment, and returns yields interpretable pathways aligned with domain knowledge, including known monetary and sectoral effects.
\end{itemize}
\noindent These results highlight the benefits of combining causal reasoning with geometric learning for trustworthy financial forecasting under non-stationary conditions.

\section{Discussion}
CSHT effectively captures dynamic, multi-hop dependencies in financial markets through a causal hypergraph framework. Unlike prior models based on shallow correlations or opaque attention, CSHT yields semantically coherent influence paths that enhance both accuracy and interpretability. Case studies confirm that its attention aligns with economic reasoning during volatile events, supporting applications in trading, risk assessment, and portfolio optimization. The model remains robust to regime shifts without retraining and is adaptable to domains such as cryptocurrencies or ESG-based investing. However, CSHT infers causal structure via attention weights, which do not imply interventional causality. Future work will incorporate counterfactual modules, such as do-calculus or invariant learning, to enable stronger causal claims.

\section{Conclusion}
We have presented CSHT, a novel Causal Sphere Hypergraph Transformer that integrates geometric learning, temporal attention, and causally structured representations to predict asset returns in dynamic financial environments. Through extensive empirical evaluations and real-world case studies, we demonstrate that CSHT delivers competitive performance, strong regime robustness, and interpretable causal attributions. Our results provide compelling evidence that embedding domain-specific economic structure via hypergraphs and causal signal flow can yield trustworthy and effective financial AI models. Beyond empirical gains, CSHT offers practical value in improving decision transparency and fostering safer deployment in high-stakes financial contexts.
Future directions include integrating interventional reasoning, scaling to cross-market generalisation, and extending CSHT to simulate economic stress tests or facilitate human-AI dialogue in portfolio diagnostics. Our work contributes towards a more transparent and causally principled foundation for next-generation financial machine learning systems.

\section{Acknowledgments}
We thank Dr. Xiaowen Dong from the University of Oxford for his insightful feedback, which helped improve this work.

% \section{Appendices}

% If your work needs an appendix, add it before the
% ``\verb|\end{document}|'' command at the conclusion of your source
% document.

% Start the appendix with the ``\verb|appendix|'' command:
% \begin{verbatim}
%   \appendix
% \end{verbatim}
% and note that in the appendix, sections are lettered, not
% numbered. This document has two appendices, demonstrating the section
% and subsection identification method.

% \section{Multi-language papers}


\bibliographystyle{ACM-Reference-Format}
\bibliography{sample-base}

\end{document}